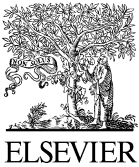

**Procedia**
Computer Science

The 14th International Conference on Ambient Systems, Networks and Technologies (ANT)
March 15-17, 2023, Leuven, Belgium

# HealthEdge: A Machine Learning-Based Smart Healthcare Framework for Prediction of Type 2 Diabetes in an Integrated IoT, Edge, and Cloud Computing System

Alain Hennebelle[a], Huned Materwala[b,c], Leila Ismail[b,c,d,*]

[a]*Independent Researcher, Melbourne, Australia*
[b]*Intelligent Distributed Computing and Systems (INDUCE) Research Laboratory, Department of Computer Science and Software Engineering, College of Information Technology, United Arab Emirates University, United Arab Emirates*
[c]*National Water and Energy Center, United Arab Emirates University, United Arab Emirates*
[d]*Cloud Computing and Distributed Systems (CLOUDS) Lab, School of Computing and Information Systems, The University of Melbourne, Australia*

**Abstract**

Diabetes Mellitus has no permanent cure to date and is one of the leading causes of death globally. The alarming increase in diabetes calls for the need to take precautionary measures to avoid/predict the occurrence of diabetes. This paper proposes HealthEdge, a machine learning-based smart healthcare framework for type 2 diabetes prediction in an integrated IoT-edge-cloud computing system. Numerical experiments and comparative analysis were carried out between the two most used machine learning algorithms in the literature, Random Forest (RF) and Logistic Regression (LR), using two real-life diabetes datasets. The results show that RF predicts diabetes with 6% more accuracy on average compared to LR.

* Corresponding author. Tel.: +971-3-7673333; fax: +971-3-7134343.
*E-mail address:* leila@uaeu.ac.ae

## 1. Introduction

Diabetes, one of the top 10 causes of death across the world [1], is a disease characterized by increased blood sugar levels [2]. Based on a report by the International Diabetes Federation, in 2021, 537 million adults globally were suffering from diabetes causing 6.7 million deaths [3]. Furthermore, the number of diabetics is projected to reach 643 million by 2030 and 783 million by 2045 [3]. Diabetes in an individual prevails due to a dynamic interaction between different risk factors such as sleep duration, alcohol consumption, dyslipidemia, physical inactivity, serum uric acid, obesity, hypertension, cardiovascular disease, family history of diabetes, ethnicity, depression, age, and gender [4]. If not treated at an early stage, diabetes can lead to severe complications [5].

The use of machine learning has thus gained wide attention for the prediction of diabetes based on risk factors data [6–13] in context of smart healthcare [14,15]. However, these works focus on stand-alone diabetes prediction. To the best of our knowledge, no work proposes a smart healthcare framework for diabetes prediction. This paper aims to address this void by proposing HealthEdge, a machine learning-based smart healthcare framework for the prediction of type 2 diabetes in an integrated IoT-edge-cloud computing system. The proposed system analyzes diabetes risk factors using medical sensors/devices and predicts the incidence of diabetes in an individual. The machine learning model is trained in the cloud and then the developed model is used by edge servers for diabetes prediction. The main contributions of this paper are as follows.

- We propose HealthEdge, a machine learning-based smart healthcare framework for diabetes prediction in an integrated IoT-edge-cloud computing system.
- We present implementation steps for the proposed system.
- The performance of the proposed system is evaluated using the two most used machine learning algorithms in literature for two real-life diabetes datasets.

Section 2 summarizes the related work on machine learning-based diabetes prediction. The proposed HealthEdge framework is explained in Section 3. Numerical experiments and comparative performance results are provided in Section 4. Finally, Section 5 concludes the paper with future research directions.

## 2. Related Work

Table 1. Related work on diabetes prediction.

| Work | Dataset | Algorithms | Evaluation metrics |
|------|---------|-----------|--------------------|
| [6] | Private (from hospitals in Saudi Arabia) | LR, SVM, DT, RF, and EMV | Accuracy, precision, recall, and F-measure |
| [7] | Private (from Slovenian primary healthcare institutions) | Linear regression, Glmnet, RF, XGBoost, and light GBM | AUC and RMSE |
| [8] | Private (from a private medical institute, Hanaro Medical Foundation, in Seoul, South Korea) | LR, RF, SVM, XGBoost[†], stacking[†], soft voting[†], and confusion matrix-based ensemble[†] | Accuracy, precision, recall, F-measure, MCC, and KC |
| [9] | **Dataset 1**: cross-sectional diabetes survey in Saudi Arabia, **Dataset 2**: NHANES, **Dataset 3**: PIMA Indian | BPM, AP, DF, LD-SVM, DJ, boosted DT, and NN | Accuracy, precision, recall, F-measure, and AUC |
| [10] | PIMA Indian | LR and DT | Accuracy, error rate, AIC, BIC, R2, and LL |
| [11] | PIMA Indian | NB, RF, and DT | Accuracy, precision, sensitivity, specificity, F-measure, and AUC |
| [12] | Henan rural cohort study | LR, CART, ANN, SVM, RF, and GBM | AUC, sensitivity, specificity, positive and negative prediction values, and area under the precision-recall curve |
| [13] | CBHS health funds company in Australia | LR, kNN, SVM, NM, DT, RF, XGBoost, and ANN | Accuracy, precision, recall, F-measure, and AUC |

LR – Logistic Regression; SVM – Support Vector Machine; DT – Decision Tree; RF – Random Forest; EMV – Ensemble Majority Voting (LR, SVM, and DT); Glment – Regularized Generalized Linear Model; XGBoost – Extreme Gradient Boosting; GBM – Gradient Boosting Machine; AUC – Area Under the ROC Curve; RMSE – Root Mean Squared Error; [†] – Ensemble algorithms used are LR, RF, SVM, and XGBoost; MCC – Mathews Correlation Coefficient; KC – Kappa's Coefficient; NHANES – National Health and Nutrition Examination Survey; BPM – Bayes Point Machine; AP – Average Perceptron; DF – Decision Forest; LD-SVM – Locally Deep SVM; DJ – Decision Jungle; NN – Neural Network; AIC – Akaike's Information Criteria; BIC – Bayesian Information Criteria; LL – Log Likelihood; NB – Naïve Bayes; CART – Classification and Regression Tree; kNN – k Nearest Neighbor

Table 1 summarizes the works on machine learning-based stand-alone diabetes prediction. None of these works propose a framework for diabetes prediction in an integrated IoT-edge-cloud computing system. In this paper, we address this gap.

## 3. HealthEdge: Machine Learning-based Smart Healthcare Framework for Predicting Type 2 Diabetes in an Integrated IoT-Edge-Cloud Computing System

Fig. 1 shows our proposed HealthEdge system in an integrated IoT-edge-cloud computing system. The IoT devices and sensors are used to collect diabetes risk factors data for the user. Different sensors and medical devices are used to measure the values of heart rate, hypertension, obesity, sleep duration, glucose, cholesterol level, physical activity, and serum uric acid. Considering the limited computing capabilities of these low -powered IoT devices, the collected data is sent to the edge servers for transformation via a mobile application. Edge servers transform the data into a proper structure consisting of time stamps, patient/user ID, and risk factors values. The transformed data is sent to the cloud [16–19] for machine learning model development. The data is first preprocessed to remove missing values using Algorithm 1. The preprocessed data is used for hyperparameter tuning (Algorithm 2) to find the optimal parameters and train a machine learning model. These values are then used to create a prediction model. Algorithm 3 presents the pseudocode for model building and validation. The developed model is then used by edge servers for predicting type 2 diabetes. Fig. 2 shows the test setup for HealthEdge.

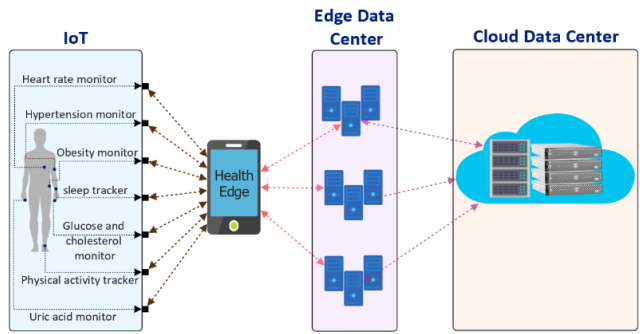
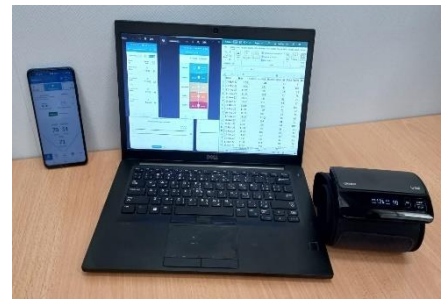

Fig. 1. Proposed HealthEdge system in an integrated IoT-edge-cloud computing system.          Fig. 2. HealthEdge test setup.

---

**Algorithm 1:** Data Preprocessor

*/* n_obs: number of observations in the dataset, n_features: number of features in the dataset                                */*
1   *Dataset* ← ReadData(*dataset.csv*)  */* reading the dataset                                */*
    */* removing observations having missing value for at least one feature                                */*
2   **for** *i from 1 to n_obs* **do**
3       **for** *j from 1 to n_features* **do**
4           **if** *cell is empty* **then**
5               *Dataset* ← RemoveObservation(*Dataset, i*)
6   **return** *Dataset*

---

**Algorithm 2:** Hyperparameter Tuner

*/* train_data*: training dataset, *test_data*: testing dataset, *hp_list*: list of hyperparameters' values, *hp_comb*: combination of values for different hyperparameters */*
1   *Dataset* ← ReadData(*dataset.csv*)  */* reading the dataset                                */*
2   *train_data, test_data* ← train_test_split(*Dataset, test_size = 0.3*)  */* splitting dataset for training and testing                                */*
3   *hp_perf* ← *[]*  */* creating an empty array to hold the performance of each hyperparameter combination                                */*
4   **for each** *hp_setting in hp_list* **do**  */* performing grid search                                */*
5       *Classifier* ← ClassificationAlgorithm(*hp_setting*)  */* creating a classifier using optimal parameters                                */*
6       *Classifer.fit(train_data*)  */* training the model by fitting the classifier to training dataset                                */*
7       *C_predict* ← *Classifier.predict(test_data*)  */* predicting diabetes class for testing dataset using the trained model                                */*
8       *Fmeasure* ← metrics.f1_score(*C_predict, test_data*)  */* computing performance metric using current combination of hyperparameters                                */*
9   *best_hp_comb* ← *hp_list[max_index(hp_perf)]*  */* finding the optimal values of hyperparameters                                */*
10   **return** *best_hp_comb*

---

**Algorithm 3:** Model Builder and Validator

1  $Dataset \leftarrow$ ReadData($dataset.csv$)  /* reading the dataset                                                                              */
  /* defining features and class labels in the dataset                                                                                            */
2  $Features \leftarrow$ Dataset.loc[ $:$ , $Dataset.columns\ !=\ 'Class\ '$]
3  $Class \leftarrow$ Dataset['$Class\ $']
4  $F\_train,\ F\_test,\ C\_train,\ C\_test \leftarrow$ train_test_split($Features,\ Class,\ test\_size = 0.3$)  /* splitting dataset for training and testing
*/
5  $Classifier \leftarrow$ ClassificationAlgorithm($best\_hp\_comb$)  /* creating a classifier using optimal parameters                         */
6  $Classifer.fit(F\_train,\ C\_train)$  /* training the model by fitting the classifier to training dataset                                    */
7  $C\_predict \leftarrow$ Classifier.predict($F\_test$)  /* predicting diabetes class for testing dataset using the trained model           */
  /* computing performance metrics using actual diabetes class and predicted diabetes class for testing dataset                                   */
8  $Accuracy \leftarrow$ metrics.accuracy_score($C\_test,\ C\_predict$)
9  $Recall \leftarrow$ metrics.recall_score($C\_test,\ C\_predict$)
10 $Precision \leftarrow$ metrics.precision_score($C\_test,\ C\_predict$)
11 $Fmeasure \leftarrow$ metrics.f1_score($C\_test,\ C\_predict$)

---

## 4. Performance Evaluation

### 4.1. Experimental Environment

We evaluate the performance of the most used two machine learning algorithms in literature for diabetes prediction, Random Forest [4] and Logistic Regression [4] (Table 1) (The code is available at: https://github.com/alain-hennebelle/HealthEdge). We use two publicly available diabetes datasets; PIMA Indian [20] and Sylhet [21]. PIMA Indian dataset is from the National Institute of Diabetes and Digestive and Kidney Diseases. It contains health information of female patients, at least 21 years old, from PIMA Indian heritage. The Sylhet dataset contains health information of patients of Sylhet Diabetes Hospital in Sylhet, Bangladesh. The information was collected using direct questionnaires and approved by a doctor. Table 2 represents the characteristics of the datasets under study.

Table 2. Characteristics of PIMA Indian and Sylhet datasets before preprocessing.

| Dataset | Number of features | Features | Number of observations in diabetic class | Number of observations in non-diabetic class | Total number of observations |
|---|---|---|---|---|---|
| PIMA Indian | 8 | Pregnancies, glucose concentration at 2 hours in an oral glucose tolerance test, diastolic blood pressure, triceps skin fold thickness, 2-hour serum insulin, Body Mass Index (BMI), diabetes pedigree function[1], age | 268 (34.9%) | 500 (65.1%) | 768 |
| Sylhet | 16 | Age, gender, polyuria[2], polydipsia[3], sudden weight loss, weakness, polyphagia[4], genital thrush[5], visual blurring, itching, irritability, delayed healing, partial paresis[6], muscle stiffness, alopecia[7], obesity | 320 (61.5%) | 200 (38.5%) | 520 |

[1]Diabetes pedigree function provides a synthesis of diabetes history in relatives and the genetic relationship of those relatives to the subject; [2]Polyuria is a condition where the body urinates more than usual and passes excessive or abnormally large amounts of urine each time you urinate; [3]Polydipsia is the feeling of extreme thirstiness; [4]Polyphagia, also known as hyperphagia, is the medical term for excessive or extreme hunger; [5]Genital thrush is a common infection caused by an overgrowth of the yeast; [6]Parital paresis involves the weakening of a muscle or group of muscles. It may also be referred to as partial or mild paralysis. Unlike paralysis, people with paresis can still move their muscles. These movements are just weaker than normal.; [7]Alpecia is an autoimmune disorder that causes your hair to come out, often in clumps the size and shape of a quarter

Figs. 3 and 4 show the PhiK ($\phi$k) correlation among features and diabetic/non-diabetic class for PIMA Indian and Sylhet datasets respectively. For the PIMA Indian dataset, the patient's glucose level, BMI, skin thickness, and age shows a correlation with the prevalence of diabetes (Fig. 3). Regarding the Sylhet dataset, polyuria and polydipsia tend to have a strong correlation with the diabetic/non-diabetic class, followed by partial paresis, gender, and sudden weight loss (Fig. 4).

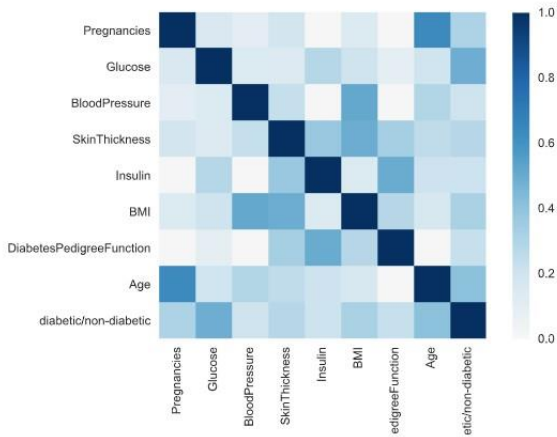

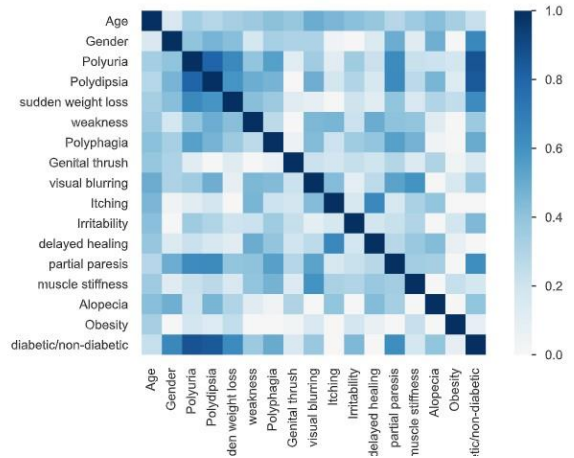

Fig. 3. Correlation between features and diabetic/non-diabetic class for PIMA Indian dataset.

Fig. 4. Correlation between features and diabetic/non-diabetic class for Sylhet dataset.

## 4.2. Experiments

We preprocess the datasets under study by removing the observations having missing values for at least one feature and normalizing values across the observations. For the PIMA Indian dataset, we remove observations having missing data for skin thickness, BMI, and blood pressure features. The Sylhet dataset has no missing values. The preprocessed PIMA Indian dataset contains 532 observations, 177 (33.3%) diabetic and 355 (66.7%) non-diabetic, and preprocessed Sylhet dataset contains 520 observations, 230 (61.5%) diabetic and 200 (38.5%) non-diabetic.

To develop an efficient diabetes prediction model for each machine learning algorithm under study, we perform hyperparameter tuning using Python's GridSearchCV library. Table 3 presents the hyperparameters for each algorithm under study and their corresponding ranges used in literature and our experiments. Furthermore, we performed feature selection and data balancing using Recursive Feature Elimination, Cross-Validated (RFECV) [22] and Synthetic Minority Oversampling Technique (SMOTE) [23] respectively. The selection of feature selection and balancing techniques are based on their efficient performances [23]. We used RFECV with Random Forest as a cross-validation evaluator. For each dataset, we use SMOTE to oversample 70% (training/validation) of the dataset, and the remaining 30% (testing) dataset is left untreated.

Table 3. Value(s) of hyperparameters used in literature and our experiments for the algorithms under study.

| Algorithm | Hyperparameters | Value(s) used in literature | Value(s) used in our experiments |
|---|---|---|---|
| | Number of estimators/trees | 100 [6], 100, 300, 500, 1000 [12], and NR [7–11,13] | 10, 20, 40, 60, 80, 100, 200, 300, 400, 500, 600, 700, 800, 900, 1000 |
| Random Forest | Maximum depth | NR [6–13] | None, 2, 5, 8 |
| | Splitting criteria | NR [6–13] | Gini and entropy |
| | Maximum features | NR [6–13] | Nmax**, sqrt, and log |
| Logistic Regression | Regularization parameter | NR [6–13] | $2^{-6}$, $2^{-4}$, $2^{-2}$, $2^{0}$, $2^{2}$, $2^{4}$, and $2^{6}$ |
| | Solver | NR [6–13] | Newton-cg, lbfgs, liblinear, sag, and saga |
| | Maximum iterations | NR [6–13] | 3000 |

NR – Not Reported; **Nmax – total number of features in a dataset; Newton-cg – Newton Conjugate Gradient

We evaluate Random Forest and Logistic Regression algorithms with and without feature selection, before and after data balancing. We use 10-fold cross-validation for testing the machine learning models under study. The performance of each algorithm for PIMA Indian and Sylhet datasets is evaluated in terms of accuracy (Equation 1), recall (Equation 2), precision (Equation 3), F-measure (Equation 4), and Area under the Curve (AUC).

$$Accuracy = \frac{TP + TN}{TP + TN + FP + FN} \quad (1)$$

$$Precision = \frac{TP}{TP + FP} \quad (3)$$

$$Recall = \frac{TP}{TP + FN} \quad (2)$$

$$F - measure = \frac{2(recall \times precision)}{recall + precision} \quad (4)$$

where TP (True Positive) represents the number of correctly classified diabetic observations, TN (True Negative) represents the number of correctly classified non-diabetic observations, FP (False Positive) represents the number of incorrectly classified non-diabetic observations, and FN (False Negative) represents the number of incorrectly classified diabetic observations.

### 4.3. Experimental Results Analysis

Table 4 shows the optimal values of hyperparameters obtained for the algorithms under study after parameter tuning. Fig. 5 and 6 show the result of feature selection for PIMA Indian and Sylhet datasets respectively. Fig. 7 and 8 show the Accuracy, F-measure, Recall, Precision, and AUC values of the models for the different datasets, with and without feature selection before and after balancing for PIMA Indian and Sylhet datasets respectively. From the raw results of accuracy, we can see that feature selection improves or at least does not degrade accuracy. As for balancing, we observe mixed results on accuracy depending on how balanced the datasets are initially. Accuracy is improved with data balancing in the case of RF algorithm with an increase from 0.77 to 0.81 for PIMA Indian and 0.97 to 0.98 for Sylhet.

The datasets we used are not heavily imbalanced, so we can notice that balancing does not show accrued accuracy (except for Sylhet dataset where there is a slight improvement). The balancing algorithm (SMOTE) is inefficient in producing better training in this case, this is because SMOTE oversamples uninformative samples. As a general result, we can observe that feature selection does not degrade the accuracy and can reduce the processing time. The Best ML algorithm for PIMA Indian dataset is Random Forest when using Feature Selection with an accuracy score of 0.7827 The Best ML algorithm for Sylhet is Random Forest with an accuracy score of 0.9723, the feature selection brings a slight decrease in processing time and the accuracy does not suffer. The results for Sylhet datasets are exceedingly better than for PIMA Indian. The main difference between these datasets is the number of available features that can be seen as risk factors for diabetes. In conclusion, we can say that we should strive to get data with as many risk factors as possible (i.e., Sylhet dataset). The datasets we used for analysis were not diverse enough to assess the need for balancing the data.

Table 4. Optimal values for hyperparameters obtained in our experiments.

| Experiment | Random Forest | | | | Logistic Regression | |
|---|---|---|---|---|---|---|
| | Number of estimators/trees | Maximum depth | Splitting criteria | Maximum features | Regularization parameter | Solver |
| PIMA Indian – No FS and No Bal | 50 | 5 | Entropy | None | $2^4$ | lbfgs |
| PIMA Indian – FS and No Bal | 40 | 5 | Entropy | Sqrt | $2^4$ | liblinear |
| PIMA Indian – FS and Bal | 50 | None | Entropy | Sqrt | $2^{-2}$ | liblinear |
| PIMA Indian – No FS and Bal | 50 | 8 | Entropy | Log2 | $2^2$ | lbfgs |
| Sylhet – No FS and No Bal | 20 | None | Gini | Sqrt | $2^4$ | lbfgs |
| Sylhet – FS and No Bal | 100 | None | Gini | Sqrt | $2^2$ | lbfgs |
| Sylhet – FS and Bal | 50 | None | Gini | Sqrt | $2^2$ | lbfgs |
| Sylhet – No FS and Bal | 50 | None | Entropy | Sqrt | $2^2$ | lbfgs |

FS – Feature Selection; Bal – Balancing

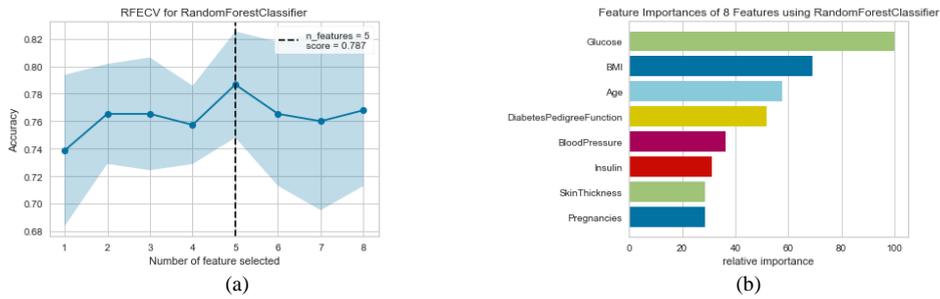

Fig. 5. Performance of feature selection for PIMA Indian dataset (a) RFECV performance; (b) Importance of features.

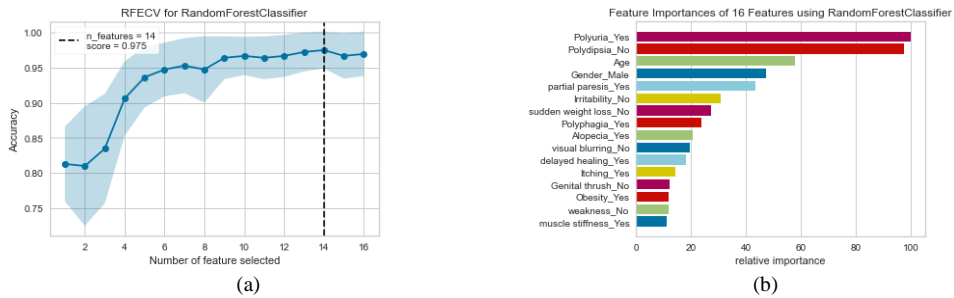

Fig. 6. Performance of feature selection for Sylhet dataset (a) RFECV performance; (b) Importance of features.

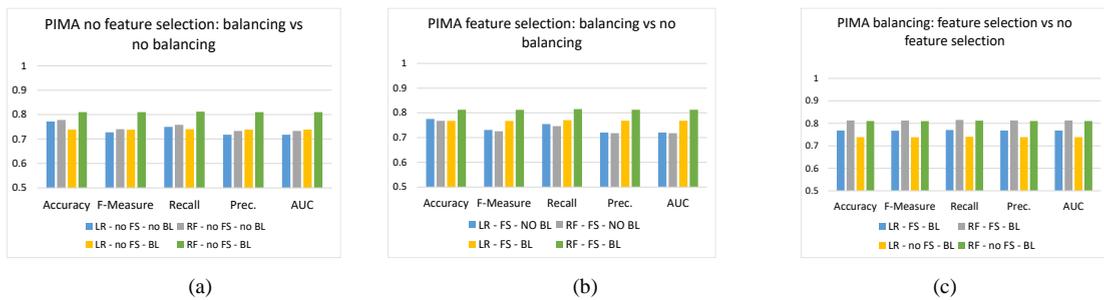

Fig. 7. Performance of algorithms under study using PIMA Indian dataset (a) No feature selection: Balancing vs no balancing; (b) Feature selection: Balancing vs no balancing; (c) Balancing: feature selection vs no feature selection.

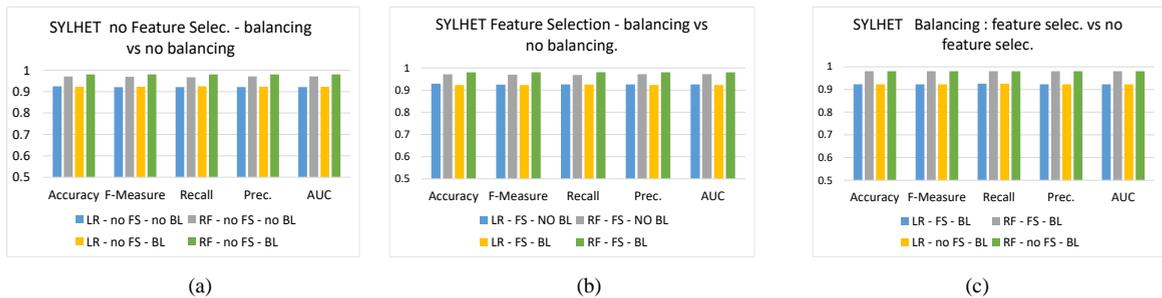

Fig. 8. Performance of algorithms under study using Sylhet dataset (a) No feature selection: Balancing vs no balancing; (b) Feature selection: Balancing vs no balancing; (c) Balancing: feature selection vs no feature selection.

## 5. Conclusions

In this paper, we propose HeatlhEdge, a machine learning-based diabetes prediction framework using IoT devices in an integrated edge-cloud computing system. In addition, we evaluate the framework using the mostly used machine learning algorithms and real-life diabetes datasets. We provide insights on risk factors that are significant for prevalence of diabetes. For future work, a large spectrum of machine learning and deep learning algorithms will be considered to evaluate the proposed system.

## Acknowledgements

This research was funded by the National Water and Energy Center of the United Arab Emirates University (Grant 12R126).

## References

[1] World Health Organization. Top 10 causes of death globally 2020. https://www.who.int/news-room/fact-sheets/detail/the-top-10-causes-of-death#:~:text=The%20top%20global%20causes%20of,birth%20asphyxia%20and%20birth%20trauma%2C (accessed November 12, 2022).

[2] Ismail L, Materwala H, al Kaabi J. Association of risk factors with type 2 diabetes: A systematic review. Comput Struct Biotechnol J 2021;19:1759–85. https://doi.org/10.1016/j.csbj.2021.03.003.

[3] International Diabetes Federation. Diabetes around the world in 2021 2022. https://diabetesatlas.org/#:~:text=Diabetes around the world in 2021%3A a- and middle-income countries. (accessed June 26, 2022).

[4] Ismail L, Materwala H, Tayefi M, Ngo P, Karduck AP. Type 2 Diabetes with Artificial Intelligence Machine Learning: Methods and Evaluation. Archives of Computational Methods in Engineering 2022;29:313–333. https://doi.org/10.1007/s11831-021-09582-x.

[5] Ismail L, Materwala H. IDMPF: intelligent diabetes mellitus prediction framework using machine learning. Applied Computing and Informatics 2021. https://doi.org/10.1108/ACI-10-2020-0094.

[6] Ahmad HF, Mukhtar H, Alaqail H, Seliaman M, Alhumam A. Investigating health-related features and their impact on the prediction of diabetes using machine learning. Applied Sciences (Switzerland) 2021;11:1–18. https://doi.org/10.3390/app11031173.

[7] Kopitar L, Kocbek P, Cilar L, Sheikh A, Stiglic G. Early detection of type 2 diabetes mellitus using machine learning-based prediction models. Sci Rep 2020;10:1–12. https://doi.org/10.1038/s41598-020-68771-z.

[8] Deberneh HM, Kim I. Prediction of type 2 diabetes based on machine learning algorithm. Int J Environ Res Public Health 2021;18:9–11. https://doi.org/10.3390/ijerph18063317.

[9] Syed AH, Khan T. Machine learning-based application for predicting risk of type 2 diabetes mellitus (t2dm) in saudi arabia: A retrospective cross-sectional study. IEEE Access 2020;8:199539–61. https://doi.org/10.1109/ACCESS.2020.3035026.

[10] Joshi RD, Dhakal CK. Predicting type 2 diabetes using logistic regression and machine learning approaches. Int J Environ Res Public Health 2021;18. https://doi.org/10.3390/ijerph18147346.

[11] Chang V, Bailey J, Xu QA, Sun Z. Pima Indians diabetes mellitus classification based on machine learning (ML) algorithms. Neural Comput Appl 2022;0123456789. https://doi.org/10.1007/s00521-022-07049-z.

[12] Zhang L, Wang Y, Niu M, Wang C, Wang Z. Machine learning for characterizing risk of type 2 diabetes mellitus in a rural Chinese population: the Henan Rural Cohort Study. Sci Rep 2020;10:1–10. https://doi.org/10.1038/s41598-020-61123-x.

[13] Lu H, Uddin S, Hajati F, Moni MA, Khushi M. A patient network-based machine learning model for disease prediction: The case of type 2 diabetes mellitus. Applied Intelligence 2022;52:2411–22. https://doi.org/10.1007/s10489-021-02533-w.

[14] Ismail L, Zhang L. Information innovation technology in smart cities. 2018. https://doi.org/10.1007/978-981-10-1741-4.

[15] Ismail L, Materwala H, P. Karduck A, Adem A. Requirements of Health Data Management Systems for Biomedical Care and Research: Scoping Review. J Med Internet Res 2020;22. https://doi.org/10.2196/17508.

[16] Ismail L, Barua R. Implementation and Performance Evaluation of a Distributed Conjugate Gradient Method in a Cloud Computing Environment. Softw Pract Exp 2012.

[17] Ismail L. Dynamic Resource Allocation Mechanisms for Grid Computing Environment. 2007 3rd International Conference on Testbeds and Research Infrastructure for the Development of Networks and Communities, 2007. https://doi.org/10.1109/TRIDENTCOM.2007.4444737.

[18] Leila I, Bruce M, Alain H. A formal model of dynamic resource allocation in Grid computing environment. 2008 Ninth ACIS International Conference on Software Engineering, Artificial Intelligence, Networking, and Parallel/Distributed Computing, 2008. https://doi.org/10.1109/SNPD.2008.136.

[19] Ismail L, Hagimont D, Mossiere J. Evaluation of the mobile agents technology: Comparison with the Client/Server Paradigm. Information Science and Technology (IST) 2000;19.

[20] Smith JW, Everhart J, Dickson W, Knowler W, Johannes R. Using the ADAP learning algorithm to forecast the onset of diabetes mellitus. Proceedings of the Annual Symposium on Computer Application in Medical Care, 1988, p. 261–5.

[21] Islam MMF, Ferdousi R, Rahman S, Bushra HY. Likelihood Prediction of Diabetes at Early Stage Using Data Mining Techniques, 2020, p. 113–25. https://doi.org/10.1007/978-981-13-8798-2_12.

[22] Guyon I, Weston J, Barnhill S, Vapnik V. Gene selection for cancer classification using support vector machines. Mach Learn 2002;46:389–422.

[23] Chawla N v, Bowyer KW, Hall LO, Kegelmeyer WP. SMOTE: synthetic minority over-sampling technique. Journal of Artificial Intelligence Research 2002;16:321–57.